\documentclass{article}
\usepackage{spconf,amsmath,graphicx}
\usepackage{amssymb,amsfonts}
\usepackage{graphicx}
\usepackage{textcomp}
\usepackage{caption}
\usepackage{acronym} 
\usepackage{xcolor}
\usepackage[tight,footnotesize]{subfigure}
\DeclareMathOperator*{\argmax}{arg\,max}

\usepackage{tabularx}
\usepackage{lscape}
\usepackage{mathtools}
\usepackage{booktabs}
\usepackage{balance}
\usepackage{algpseudocode}
\usepackage{algorithm}
\newcommand{\kk}[1]{}
\usepackage[none]{hyphenat}
\def\BibTeX{{\rm B\kern-.05em{\sc i\kern-.025em b}\kern-.08em
    T\kern-.1667em\lower.7ex\hbox{E}\kern-.125emX}}
\DeclareUnicodeCharacter{2212}{-}

\title{Exploring the Interplay of Interpretability and Robustness in Deep Neural Networks: A Saliency-guided Approach}
\name{Amira Guesmi\textsuperscript{\textdagger}, Nishant Suresh Aswani\textsuperscript{\textdagger,*}, and Muhammad Shafique\textsuperscript{\textdagger} 
}
\address{    \textsuperscript{\textdagger}eBrain Lab, Division of Engineering, New York University Abu Dhabi, Abu Dhabi, UAE \\
\textsuperscript{*}Department of Computer Science and Engineering, New York University Tandon, Brooklyn, USA }
\begin{document}
%
\maketitle
\begin{abstract}

Adversarial attacks pose a significant challenge to deploying deep learning models in safety-critical applications. Maintaining model robustness while ensuring interpretability is vital for fostering trust and comprehension in these models. This study investigates the impact of Saliency-guided Training (SGT) on model robustness, a technique aimed at improving the clarity of saliency maps to deepen understanding of the model's decision-making process. Experiments were conducted on standard benchmark datasets using various deep learning architectures trained with and without SGT. Findings demonstrate that SGT enhances both model robustness and interpretability. Additionally, we propose a novel approach combining SGT with standard adversarial training to achieve even greater robustness while preserving saliency map quality.
Our strategy is grounded in the assumption that preserving salient features crucial for correctly classifying adversarial examples enhances model robustness, while masking non-relevant features improves interpretability. Our technique yields significant gains, achieving a 35\% and 20\% improvement in robustness against PGD attack with noise magnitudes of $0.2$ and $0.02$ for the MNIST and CIFAR-10 datasets, respectively, while producing high-quality saliency maps.
\end{abstract}
\section{Introduction}
\label{sec:intro}
Deep neural networks (DNNs) have demonstrated remarkable performance in addressing a multitude of complex real-life problems \cite{al2017deep, miotto2018deep, yolov7}. However, their effectiveness is often hindered by a lack of interpretability and their inherent susceptibility to adversarial attacks \cite{fgsm}. In critical domains such as healthcare, neuroscience, finance, and autonomous driving \cite{caruana2015intelligible, lipton2018mythos, weber2023applications, babaei2022explainable}, reliable explanations are essential for understanding model decisions and fostering trust in their predictions. Researchers are actively exploring ways to enhance both interpretability and robustness in deep learning models. It's an ongoing challenge to strike a balance between these two objectives, and advancements in this area are crucial for deploying deep learning models in safety-critical applications.

The relationship between interpretability and robustness against adversarial attacks in deep learning models represents a complex and evolving area of research. Examining whether more interpretable models inherently possess greater robustness is a nuanced endeavor. Some researchers assert that interpretability can facilitate the identification of vulnerabilities and provide insights into model behavior, thereby fostering the development of more robust models. Conversely, others \cite{opposite} argue that interpretability and robustness are not inherently linked, suggesting that it is possible to have models that are both interpretable and vulnerable, or conversely, models that are robust but lack interpretability. This ongoing debate underscores the multifaceted nature of the relationship between interpretability and robustness in deep learning models, necessitating further investigation and exploration.

One technique aimed at enhancing model interpretability is Saliency-guided Training (SGT), introduced by Ismail et al. \cite{sgt_ismail}, which emphasizes important features in the input data while mitigating the influence of non-relevant and shortcut features. This method enriches training samples by generating new inputs through feature masking, particularly targeting regions with low gradient values.

In this paper, we investigate the impact of saliency-guided training on model robustness. Contrary to previous findings \cite{opposite}, we demonstrate that SGT indeed enhances model resilience against strong white-box adversarial attacks. Additionally, we propose a novel training technique that combines both SGT and adversarial training (AT) to produce deep neural networks that are both robust and interpretable. This strategy is based on the assumption that salient features, which contribute to the correct classification of an adversarial example, should be preserved, while non-relevant features can be masked to both increase model robustness and enhance model interpretability.

\noindent\textbf{Novel Contributions -- } The main contributions of this paper are: 
\begin{itemize}
    \item We explore the influence of Saliency-guided Training on model robustness, revealing its ability to enhance resilience against adversarial attacks, contrary to previous findings. We provide empirical evidence supporting the effectiveness of SGT in improving model robustness. 
    \item We propose a novel training technique that integrates SGT and adversarial training (AT) (named adversarial saliency-guide training (ASGT)), aiming to produce deep neural networks that are both robust and interpretable.
    \item Our technique ASGT demonstrates substantial improvements, achieving a 35\% and 20\% enhancement in robustness under the PGD attack for a noise magnitude of $0.2$ and $0.02$  for the MNIST and CIFAR-10 datasets, respectively, while also yielding higher-quality saliency maps.
    \item \textbf{Open-Source Contribution:} for reproducible research, we release the complete source code of our technique \footnote{Once paper accepted}. 
\end{itemize}

\section{Related Work}
\label{sec:related}
In this section, we delve into related works on robust training methodologies and model interpretability. 
\subsection{Adversarial Training (AT)}

Adversarial training \cite{at_madry} stands as one of the most extensively studied defenses against adversarial attacks. Its inception can be traced back to \cite{fgsm}, where models were bolstered by integrating adversarial examples into the training dataset, enabling them to recognize and defend against such instances. However, the effectiveness of adversarial training diminishes when attackers employ different attack methods from those considered during training \cite{samangouei2018defense}. Additionally, this technique imposes significant computational demands, as the generation of adversarial examples for training requires substantial time and resources. Moreover, advancements in adversarial attack techniques, such as real-time attacks, pose additional challenges in defending against adversarial examples \cite{moosavidezfooli2017universal, guesmi2022room}. 
\subsection{Saliency-guided Training (SGT) for enhancing DNN interpretability}

Recently, there has been a surge of interest in saliency-guided training techniques, which are designed to mitigate noisy gradients used in predictions while preserving the performance of models. Saliency-guided training aids in filtering out irrelevant or noisy information from the gradients, thereby leading to more accurate predictions without compromising predictive performance. Ismail et al. \cite{sgt_ismail} introduced saliency-guided training by constructing a new input through feature masking based on low gradient values. Additionally, to augment the training loss, SGT incorporates an additional regularization term aimed at minimizing the Kullback–Leibler (KL) divergence between the new and original outputs. This regularization term ensures the model produces similar output probability distributions over labels for the original clean input $X$ and the masked sample $\tilde{X}$. The detailed steps for conducting saliency-guided training are elucidated in Algorithm \ref{alg:sgt}.
\begin{algorithm}
\caption{Saliency Guided Training (SGT)}
\begin{algorithmic}[1]
\State \textbf{Input:} Training Sample $X$, $\#$ of features to be masked $k$, attack order $p$, perturbation budget $\epsilon$, learning rate $\tau$, hyperparameter $\theta$
\State \textbf{Output:}  $f_\theta$
\For {epochs}
\For {minibatches}
\State \textit{$\#$ Create the masked sample:}
\State \textit{$1.$	Get sorted index $I$ for the gradient of output with respect to the input:}
\State $I=S(\nabla_{X} f_\theta (X))$
\State \textit{$2.$ Mask bottom $k$ features:}
\State $\tilde{X}=M_k (X,I)$
\State \textit{$\#$Compute the loss:} 
\State $L_i=L(f_{\theta_i}(X),y)+\lambda D_{KL} (f_{\theta_i} ( \tilde{X})||f_{\theta_i} (X))$
\State \textit{$\#$Update $\theta$:} 
\State $\theta_{i+1}=\theta_i-\tau\nabla_{\theta_i} L_i$
\EndFor
\EndFor
\end{algorithmic}
\label{alg:sgt}
\end{algorithm}
\begin{figure*}[!tp]
    \centering
    \includegraphics[width=\textwidth]{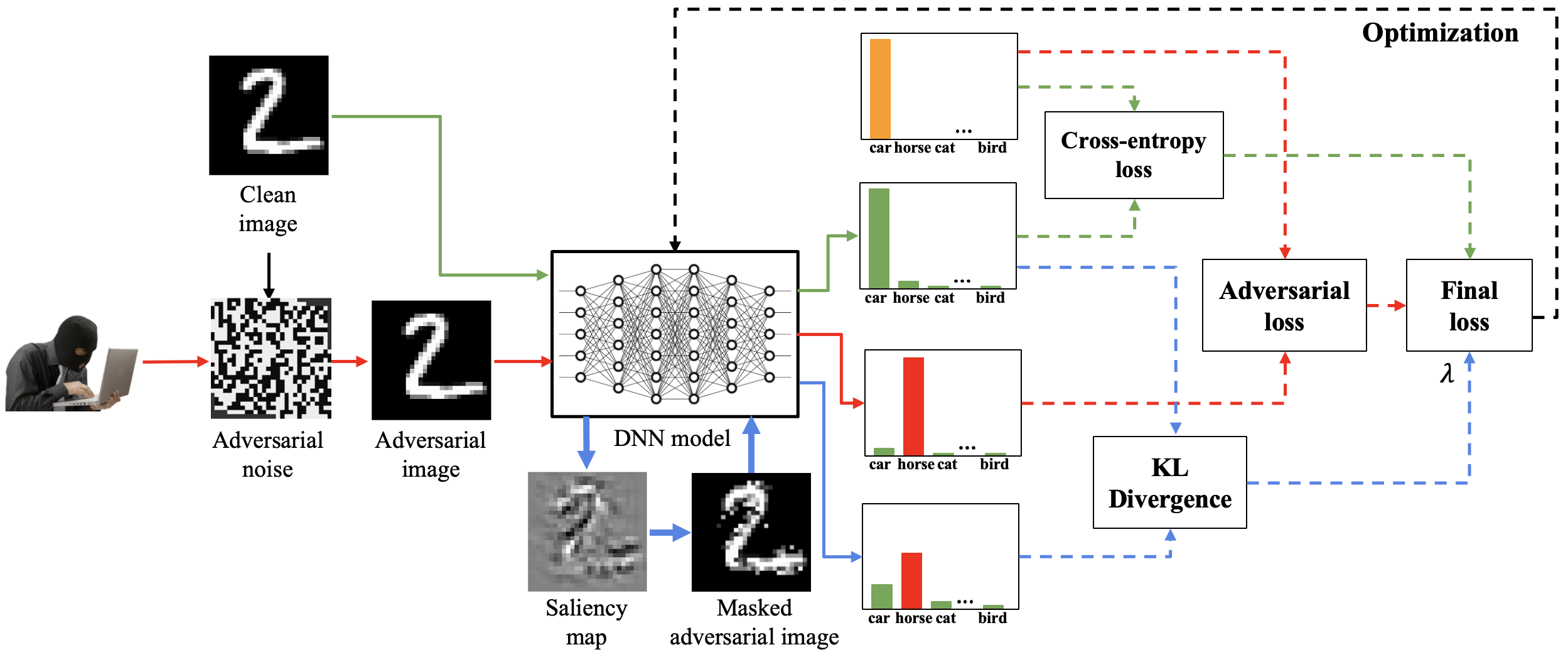}
    \caption{Overview of our proposed adversarial saliency-guided training (ASGT).}
    \label{fig:methodolgy}
\end{figure*}
\subsection{Saliency-guided Adversarial Training (SGA) for Learning Generalizable Features} %

Li et al. \cite{li2022saliency} address the decline in performance when models are tested on out-of-distribution (OOD) datasets. This issue is particularly critical in medical imaging diagnosis systems, which may demonstrate high accuracy but fail when applied to new hospitals or datasets. Recent studies suggest that these systems may learn to rely on shortcut or non-relevant features instead of relevant and generalizable ones. In \cite{li2022saliency}, authors propose a saliency guided adversarial training (SGA) to learn generalizable features. Based on the assumption that adversarial training can help eliminate shortcut features, while saliency guided training can filter out non-relevant features—both of which contribute to performance degradation on OOD test sets. Building on this hypothesis, they introduce a novel model training scheme for deep neural networks, aiming to enable the learning of good features essential for classification and/or detection tasks. 

\section{Adversarial Saliency Guided Training (ASGT)}
\label{sec:asgt}


We propose Adversarial Saliency-guided Training (ASGT), introduced in Figure \ref{fig:methodolgy}, a novel approach to train the deep neural network to learn the salient features that makes a sample behave adversarially by suppressing non-relevant and eliminating shortcut features. During ASGT, we augment the training set by generating a new adversarial training sample for each adversarial sample by masking adversarial features with low gradient values and minimizing the distance between the output of the clean sample and the masked adversarial sample (see Algorithm \ref{alg:asgt}).

In the context of a classification task on input data $\{X_i, y_i\}_{i=1}^n$, we train a deep neural network model $f_\theta$, parameterized by $\theta$, to predict the target $y$. Standard training entails minimizing the cross-entropy loss $L$ across the training set, formulated as follows:
\begin{equation}
    \min_{\theta} \frac{1}{n} \sum_{i=1}^{n} L(f_{\theta}(X_i),y_i)
\end{equation}
The model parameter $\theta$ is updated through a single step of gradient descent using the learning rate $\tau$ on a mini-batch of $m$ samples $\{(X_i, y_i)\}_{i=1}^m$:

\begin{equation}
    \theta_{i+1}=\theta_i-\tau \frac{1}{m} \sum_{i=1}^{m}\nabla_{\theta_i} L(f_{\theta}(X_i),y_i)
\end{equation}


In this work, we propose ASGT, a novel procedure to train the deep neural network models to learn the `important features' by suppressing non-relevant and further eliminating shortcut features identified in adversarial samples. 

In SGA \cite{li2022saliency}, the training set is augmented by incorporating adversarial samples generated from masked inputs. In our approach, however, we first generate the adversarial sample and subsequently apply masking to it. Our rationale behind this approach is that the salient features in the adversarial sample, which allow the model to correctly classify the adversarial example, are considered important features and should be preserved. Conversely, the rest of the features are deemed less relevant and are subject to masking.

\begin{algorithm}
\caption{Adversarial Saliency Guided Training (ASGT)}
\begin{algorithmic}[1]
\State \textbf{Input:} Training Sample $X$, $\#$ of features to be masked $k$, attack order $p$, perturbation budget $\epsilon$, learning rate $\tau$, hyperparameter $\theta$
\State \textbf{Output:}  $f_\theta$
\For {epochs}
\For {minibatches}
\State \textit{$\#$ Generate the adversarial example:}
\State $\delta^*= \argmax_{(|\delta|_p \leqslant \epsilon)} L(f_\theta (X+\delta),y),$ %
\State $X'=X+\delta^*$
\State \textit{$\#$ Create the masked adversarial example:}
\State $I=S(\nabla_{X'} f_\theta (X'))$
\State $\tilde{X}'=M_k (X',I)$
\State \textit{$\#$Compute the loss:} 
\State $L_i=L(f_{\theta_i}(X),y)+L(f_{\theta_i} (X'),y)+\lambda D_{KL} (f_{\theta_i} ( \tilde{X}')||f_{\theta_i} (X))$
\State \textit{$\#$Update $\theta$:} 
\State $\theta_{i+1}=\theta_i-\tau\nabla_{\theta_i} L_i$
\EndFor
\EndFor
\end{algorithmic}
\label{alg:asgt}
\end{algorithm}

During ASGT, we augment the training set by generating a new training sample for each adversarial sample $X'$ by masking shortcut features with low gradient values as follows:
\begin{equation}
    \tilde{X}' = M_k(X', S(\nabla_{X'} f_\theta (X')))
\end{equation}

Here, $S(.)$ represents a function that sorts the gradient of each feature from $X'$ in ascending order. $M_k(.)$ denotes an input mask function, which substitutes the $k$ lowest features from an input sample with random values within the feature range, as determined by the order specified by $S(\nabla)$. This process is based on the observation that non-relevant features typically have gradient values close to zero. The parameter $k$ serves as a tuning parameter, and its selection depends on the extent of nuisance information present in a training sample.

We aim to encourage similarity between the output of the model when provided with the masked adversarial example and when given the original clean sample. To achieve this, we use the Kullback-Leibler (KL) divergence, ensuring that the model generates similar output probability distributions across labels for both the original clean input and the masked adversarial sample.
\begin{equation}
     D_{KL} (f_{\theta_i} ( \tilde{X}')||f_{\theta_i} (X))
\end{equation}

The final loss used to update the model is defined as follows:
\begin{equation}
    L_i=L(f_{\theta_i}(X),y)+L(f_{\theta_i} (X'),y)+\lambda D_{KL} (f_{\theta_i} ( \tilde{X}')||f_{\theta_i} (X))
\end{equation}
where $\lambda$ is a hyperparameter used to weight the importance of the KL divergence relative to the other loss terms.

\section{Experiments and Results}
\label{sec:exp}

\subsection{Experimental Setup}

\begin{figure*}[!tbp]
\centering
\includegraphics[width=0.33\textwidth]{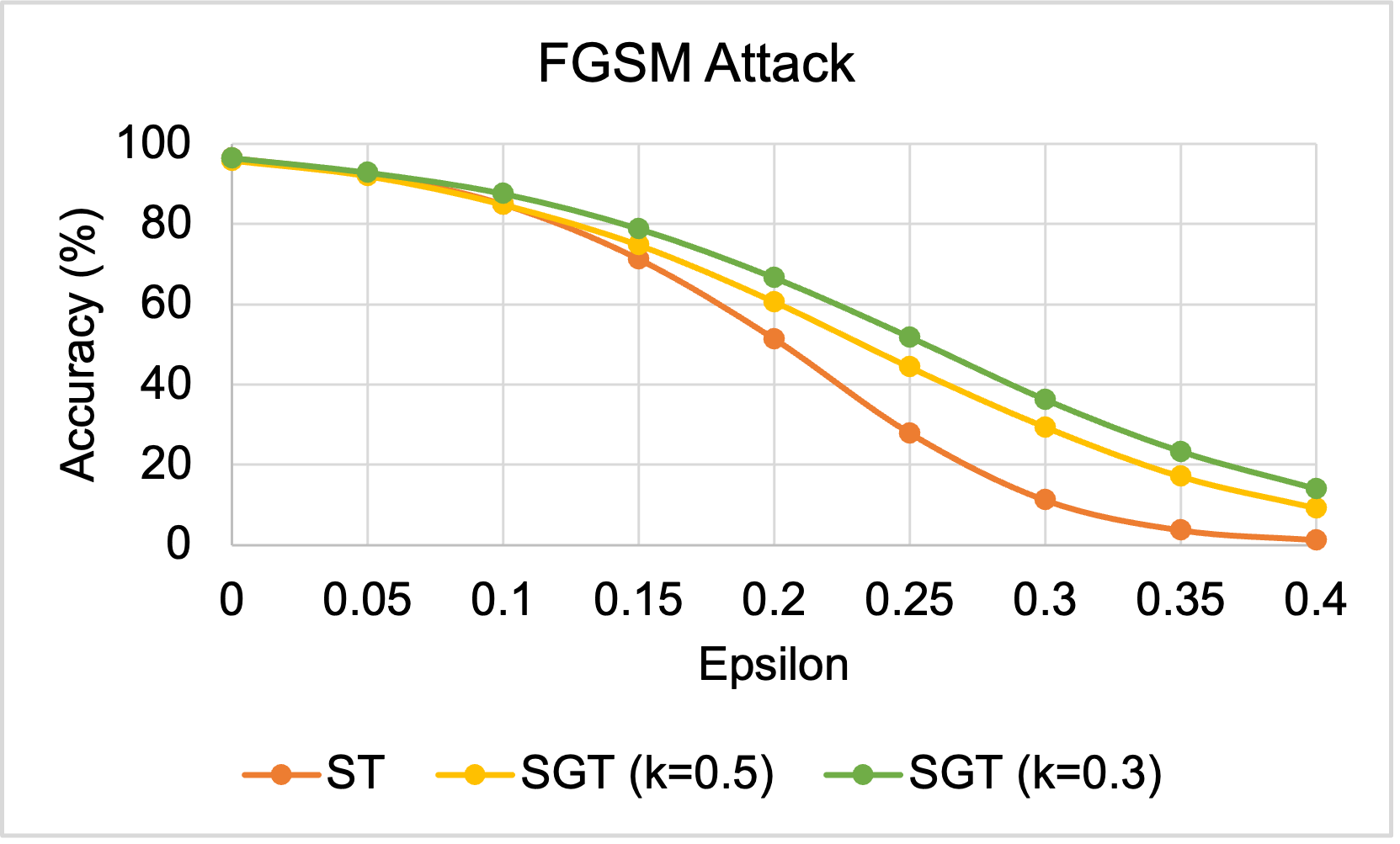}
\includegraphics[width=0.33\textwidth]{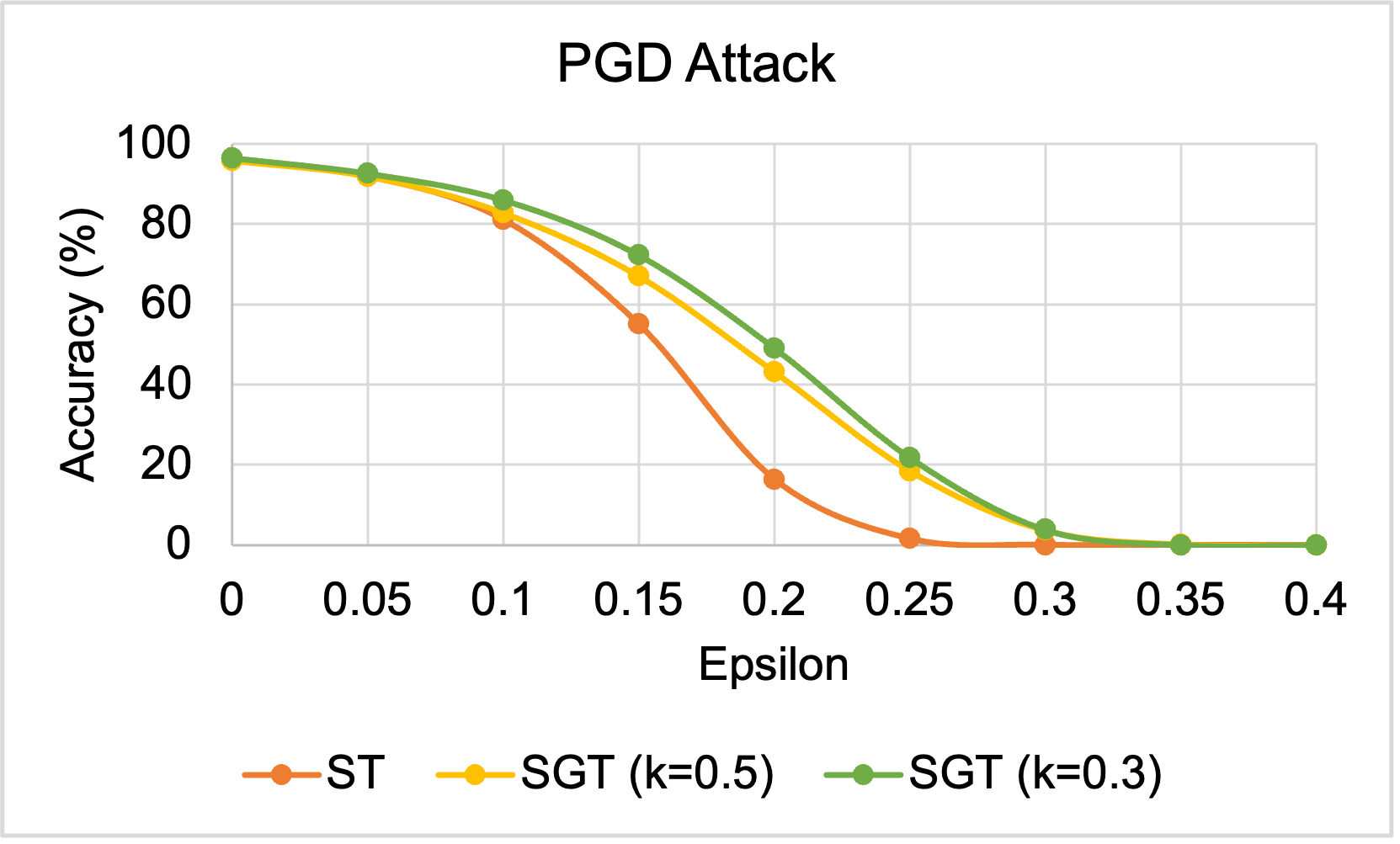}
\includegraphics[width=0.33\textwidth]{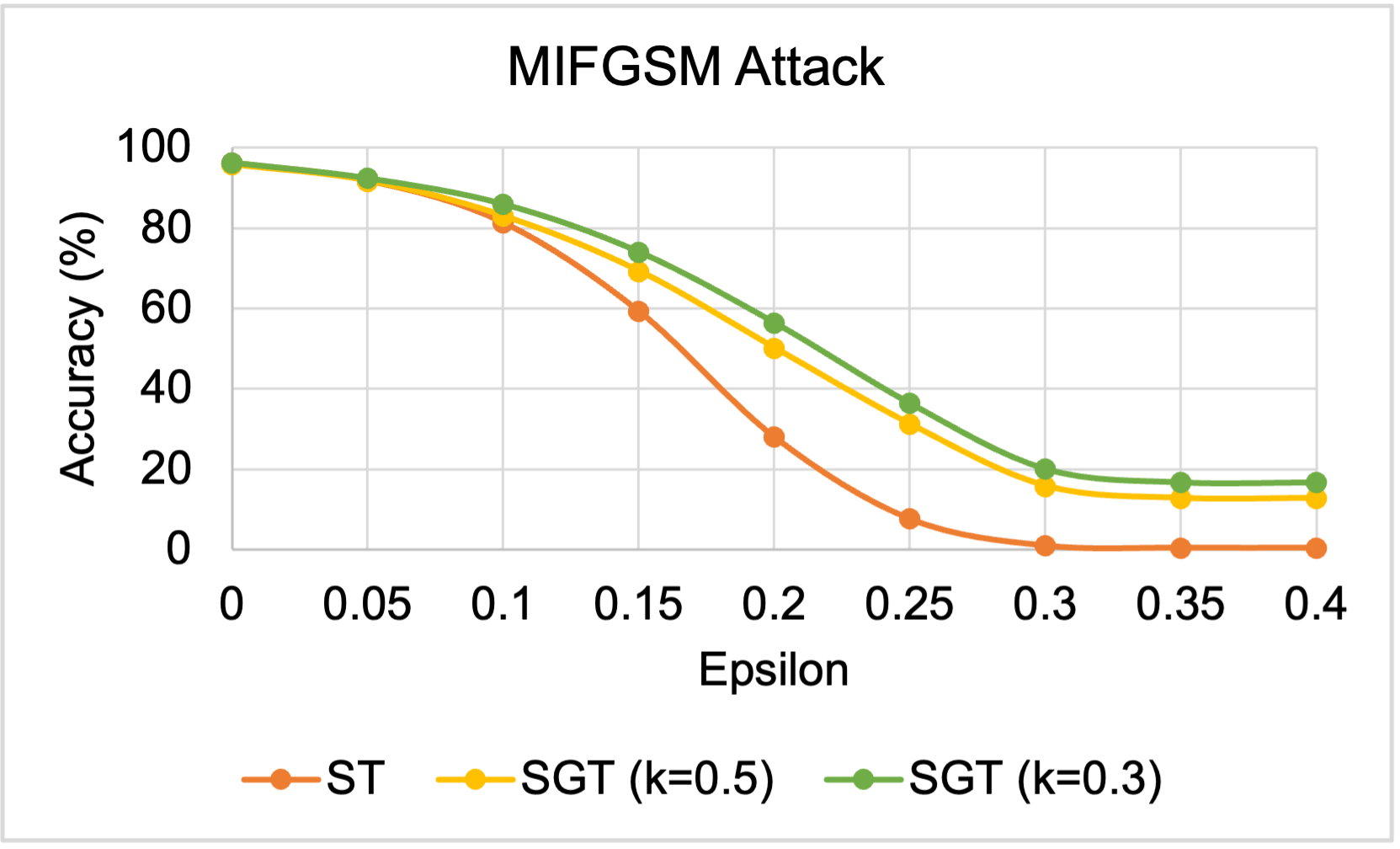}
\caption{Robustness of models against adversarial examples on the MNIST dataset. Models trained using standard training (ST) and SGT with varying degrees of feature masking (\( k = 0.3 \) and \( k = 0.5 \)) across various magnitudes of noise (\( \epsilon \)) for the FGSM, PGD, and MIFGSM attacks.}
\label{fig:sgt_mnist}
\end{figure*}

\begin{figure*}[!tbp]
\centering
\includegraphics[width=0.33\textwidth]{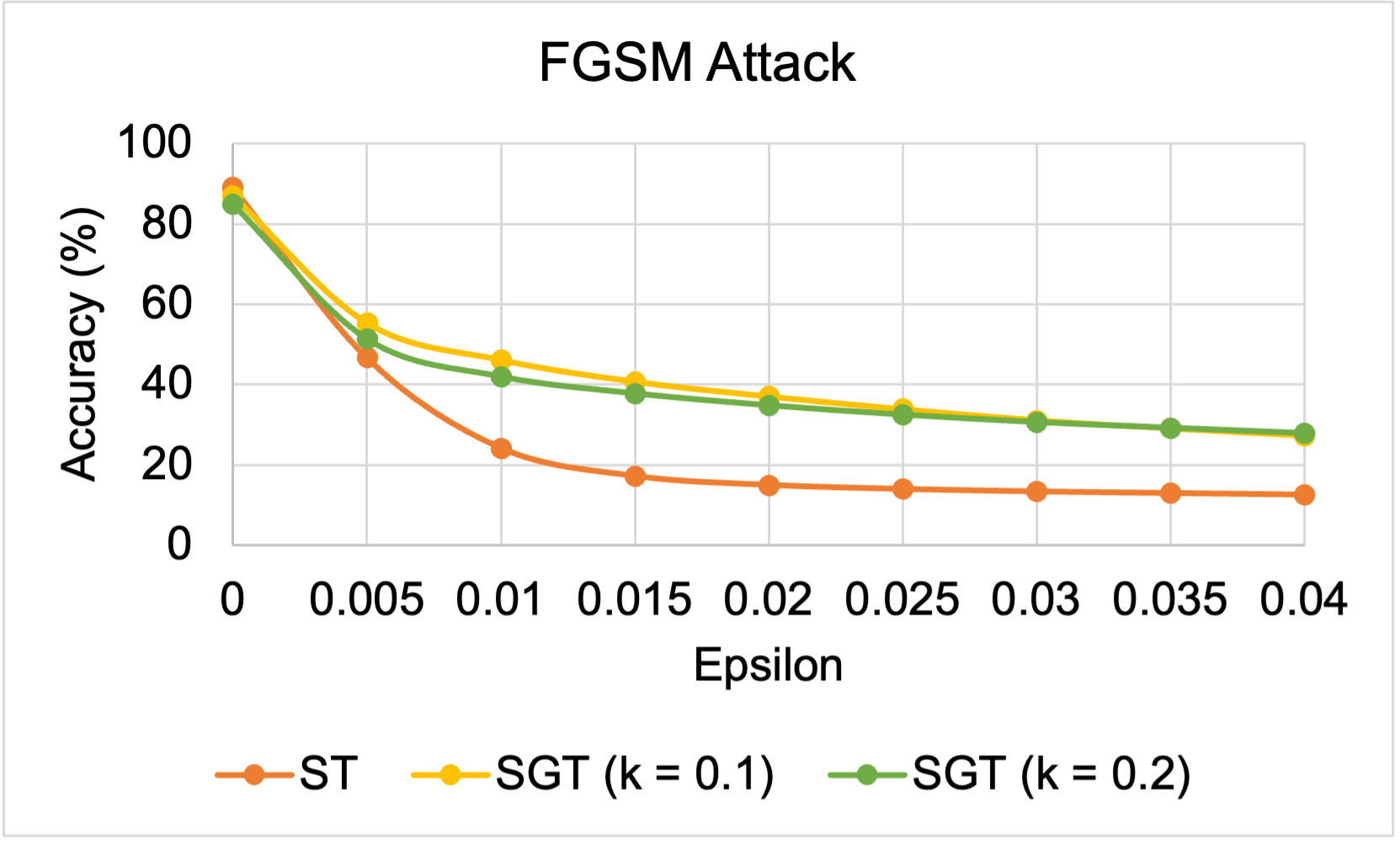}
\includegraphics[width=0.33\textwidth]{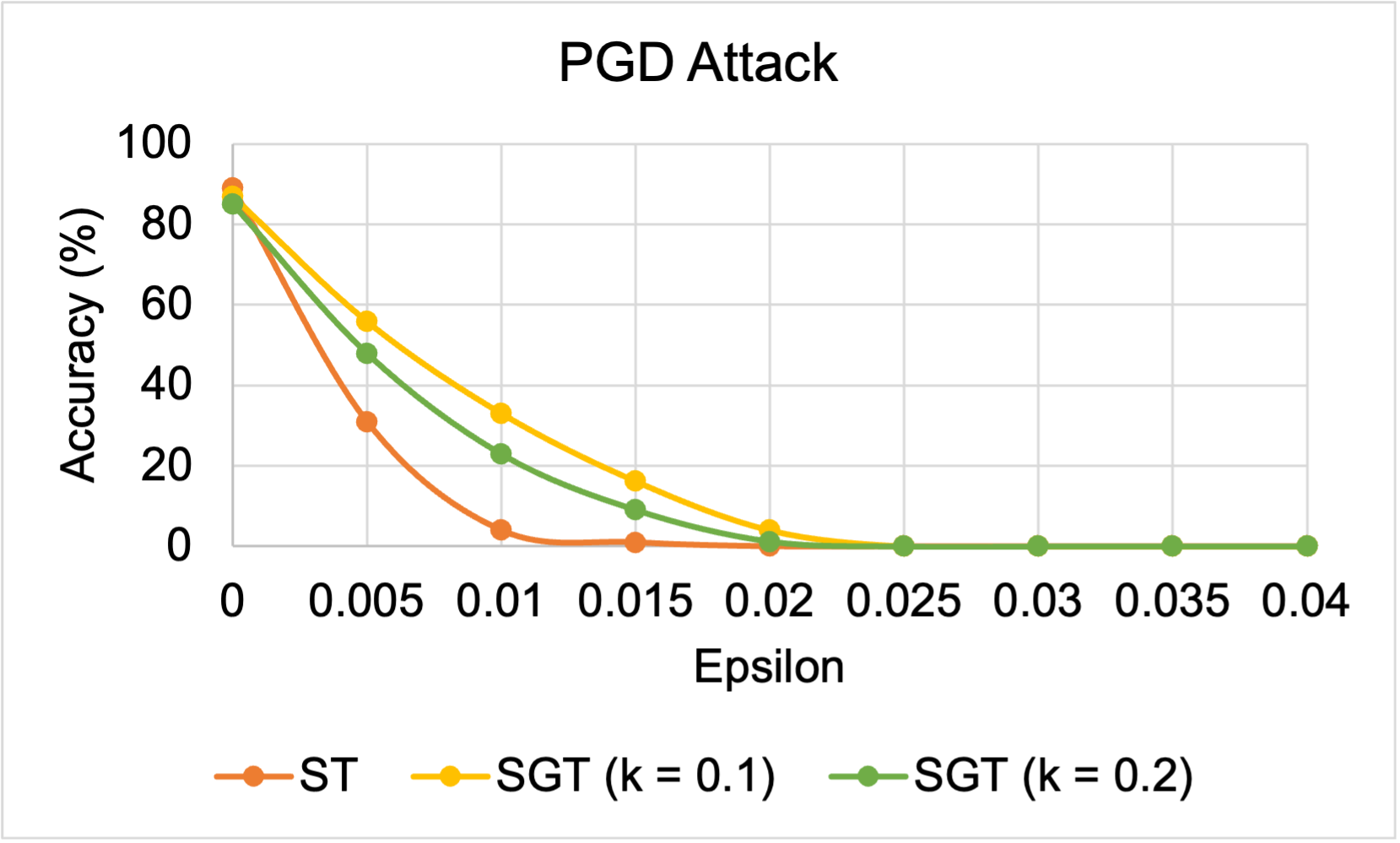}
\includegraphics[width=0.33\textwidth]{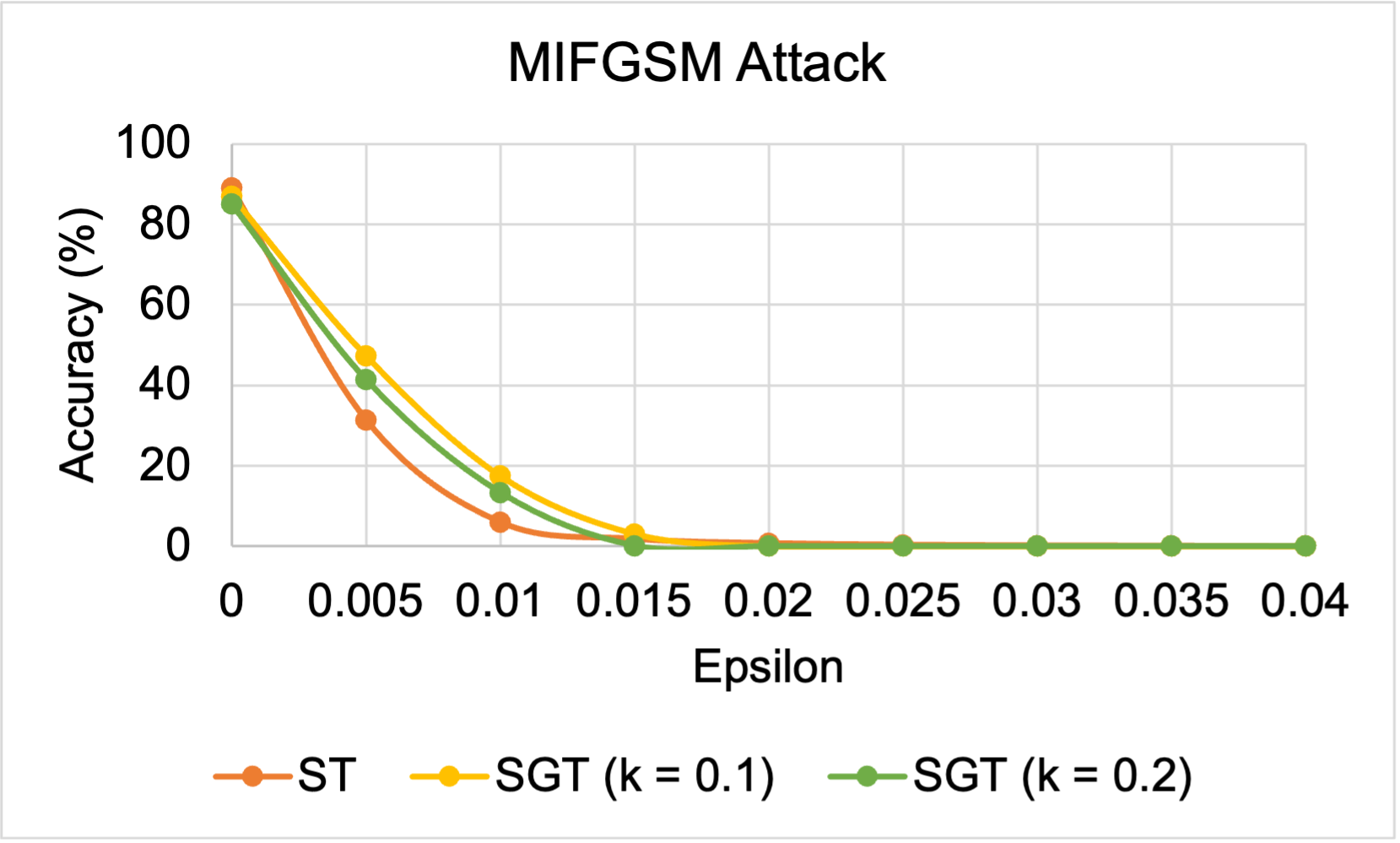}
\caption{Robustness of models against adversarial examples on the CIFAR-10 dataset. Models trained using standard training (ST) and SGT with varying degrees of feature masking (\( k = 0.1 \) and \( k = 0.2 \)) across various magnitudes of noise (\( \epsilon \)) for the FGSM, PGD, and MIFGSM attacks.}
\label{fig:sgt_cifar}
\end{figure*}

\begin{figure*}[!tbp]
\centering
\includegraphics[width=0.33\textwidth, height=3cm]{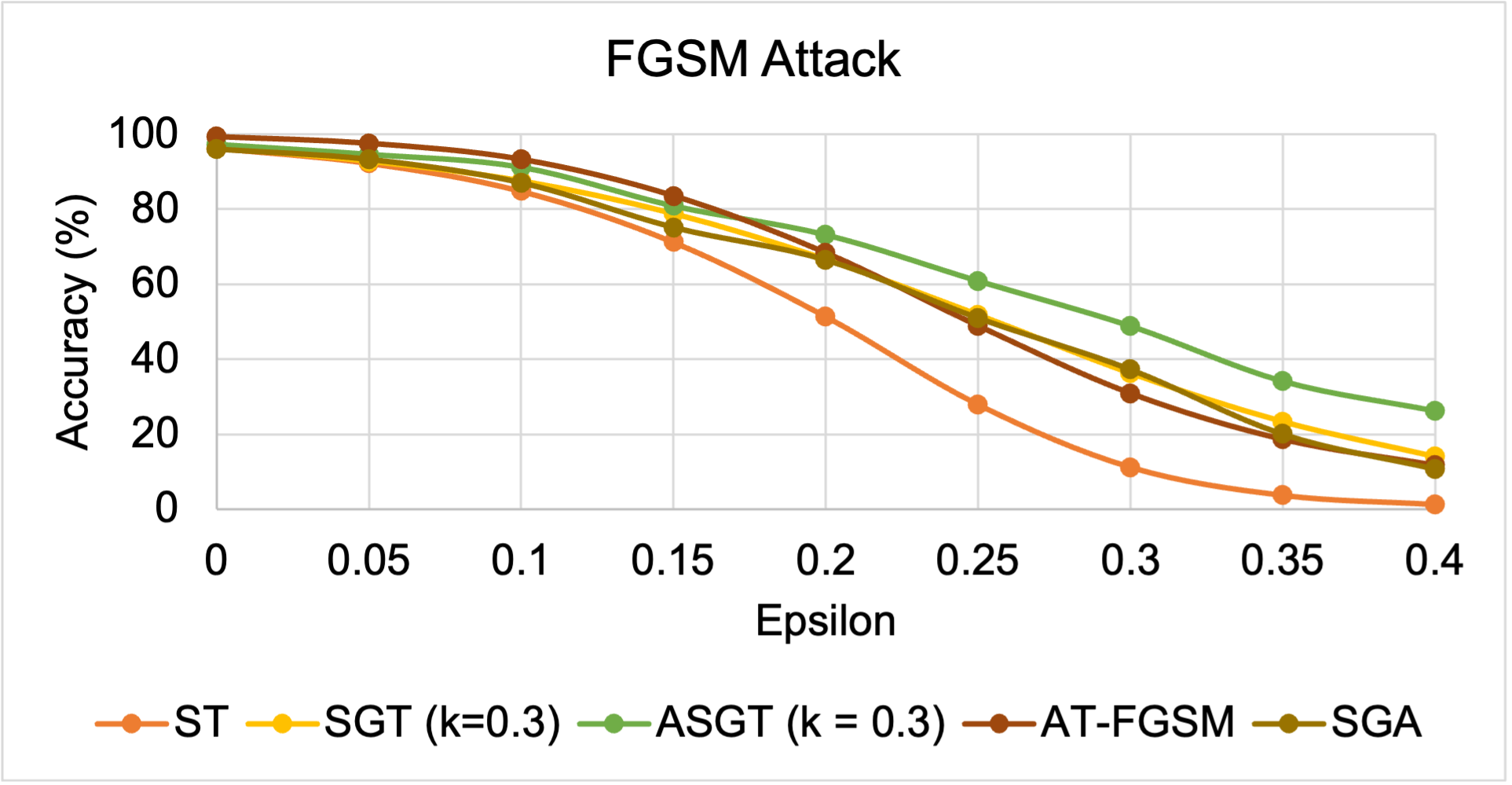}
\includegraphics[width=0.33\textwidth, height=3cm]{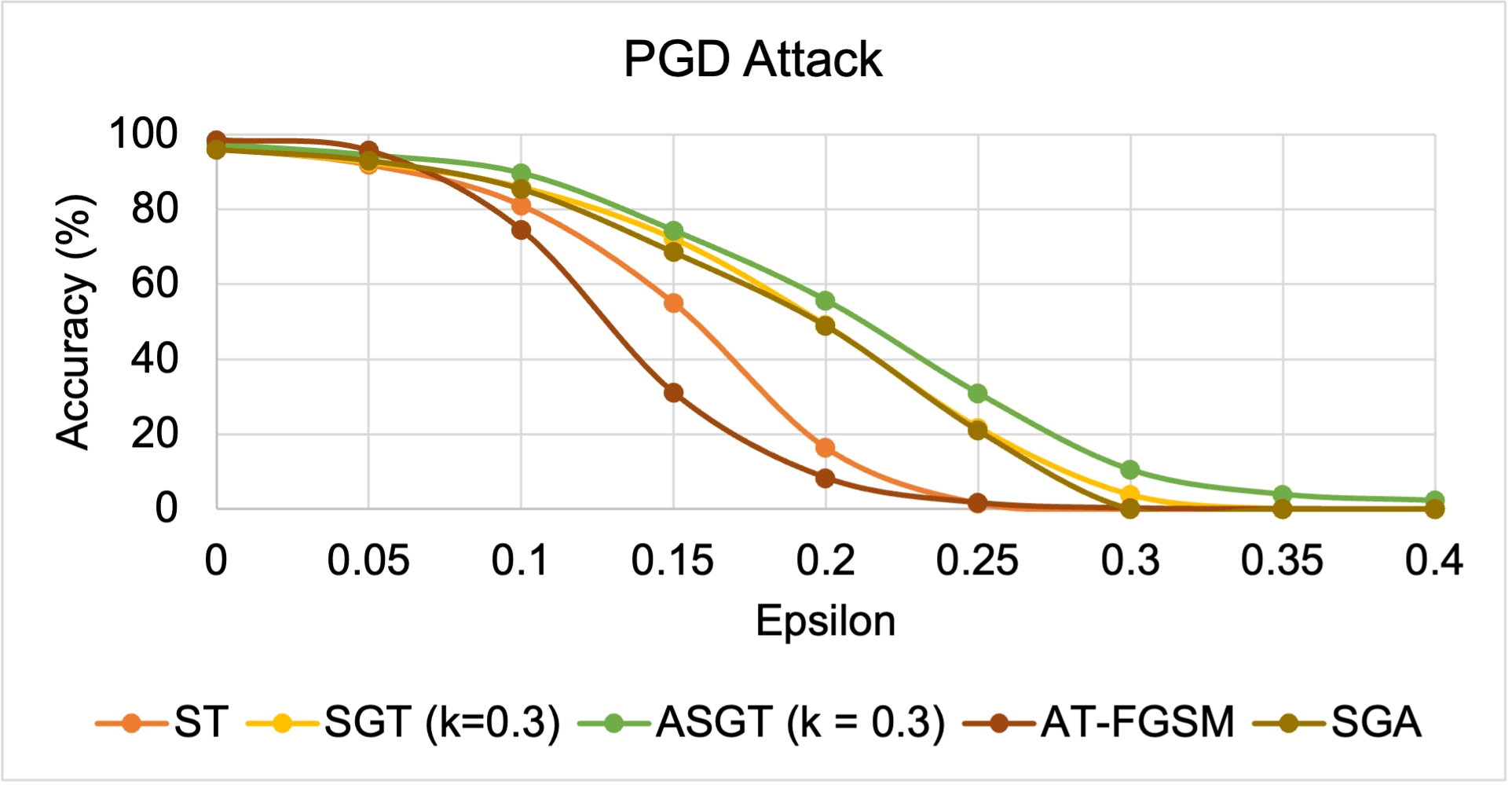}
\includegraphics[width=0.33\textwidth, height=3cm]{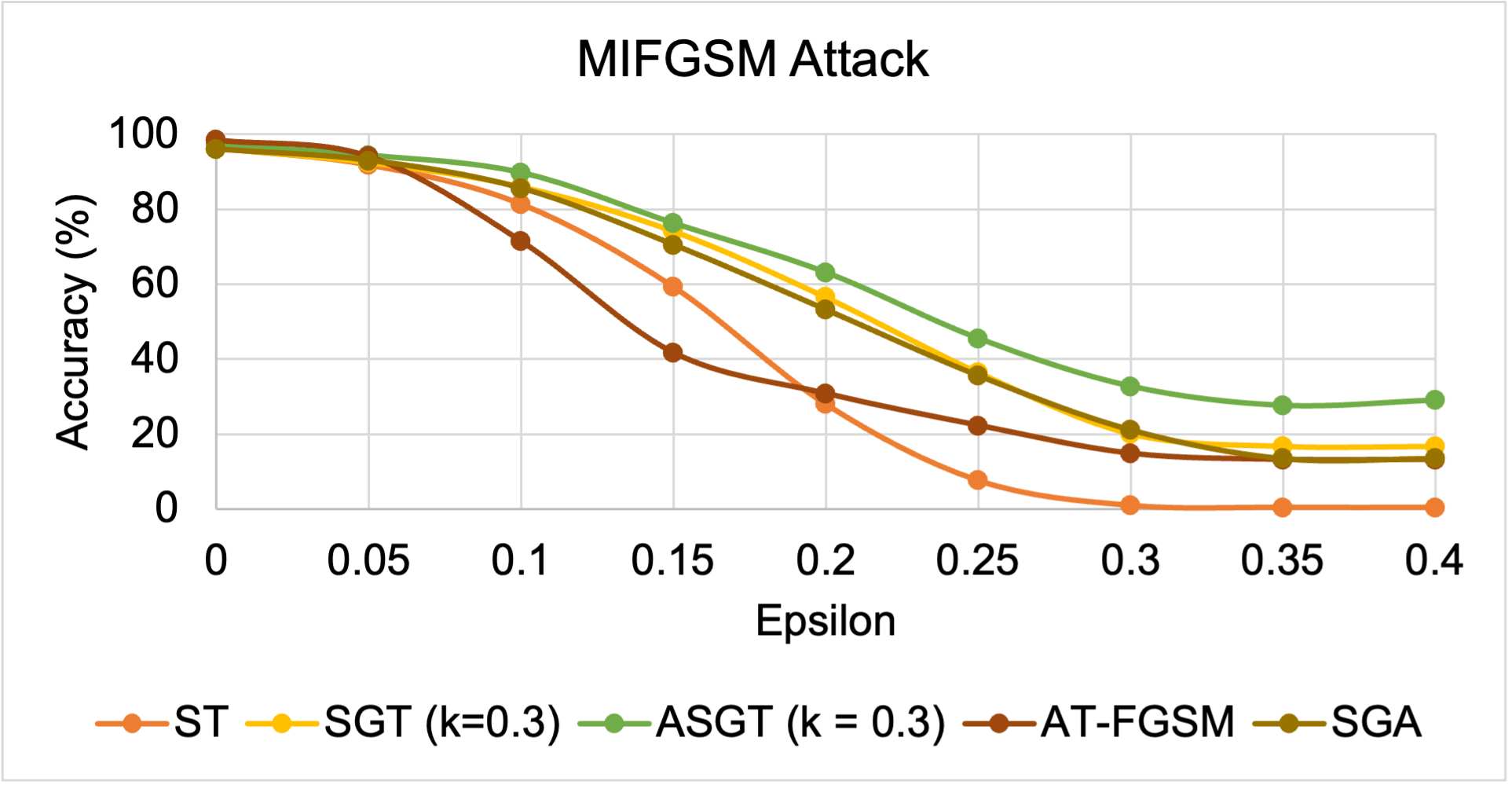}
\caption{Robustness of models against adversarial examples on the MNIST dataset. Models trained using ST, SGT (\( k = 0.3 \)), ASGT (\( k = 0.3 \)), AT-FGSM, and SGA across various magnitudes of noise (\( \epsilon \)) for the FGSM, PGD, and MIFGSM attacks.}
\label{fig:sgt_mnist_all}
\end{figure*}
\begin{figure*}[!tbp]
\centering
\includegraphics[width=0.33\textwidth, height=3cm]{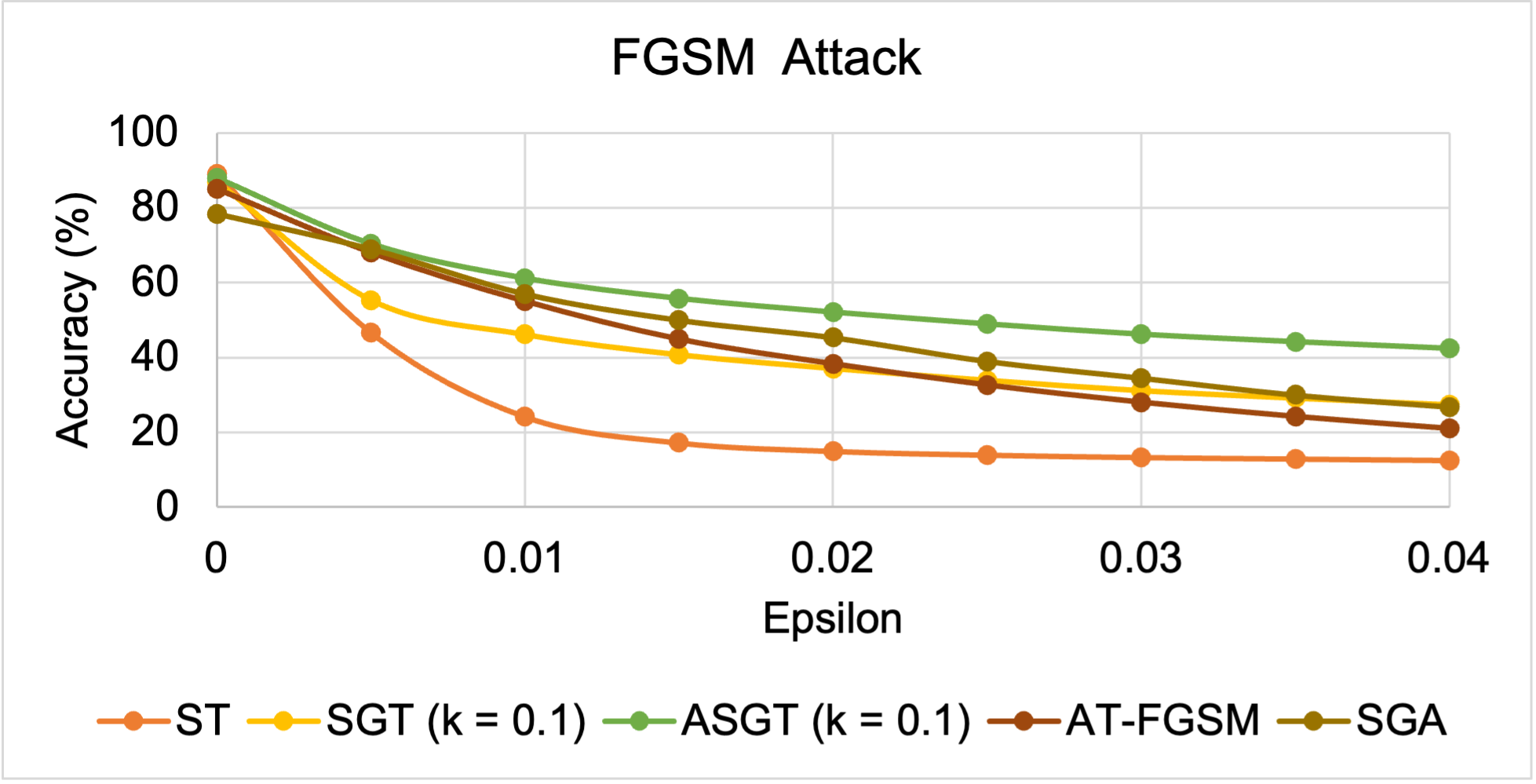}
\includegraphics[width=0.33\textwidth, height=3cm]{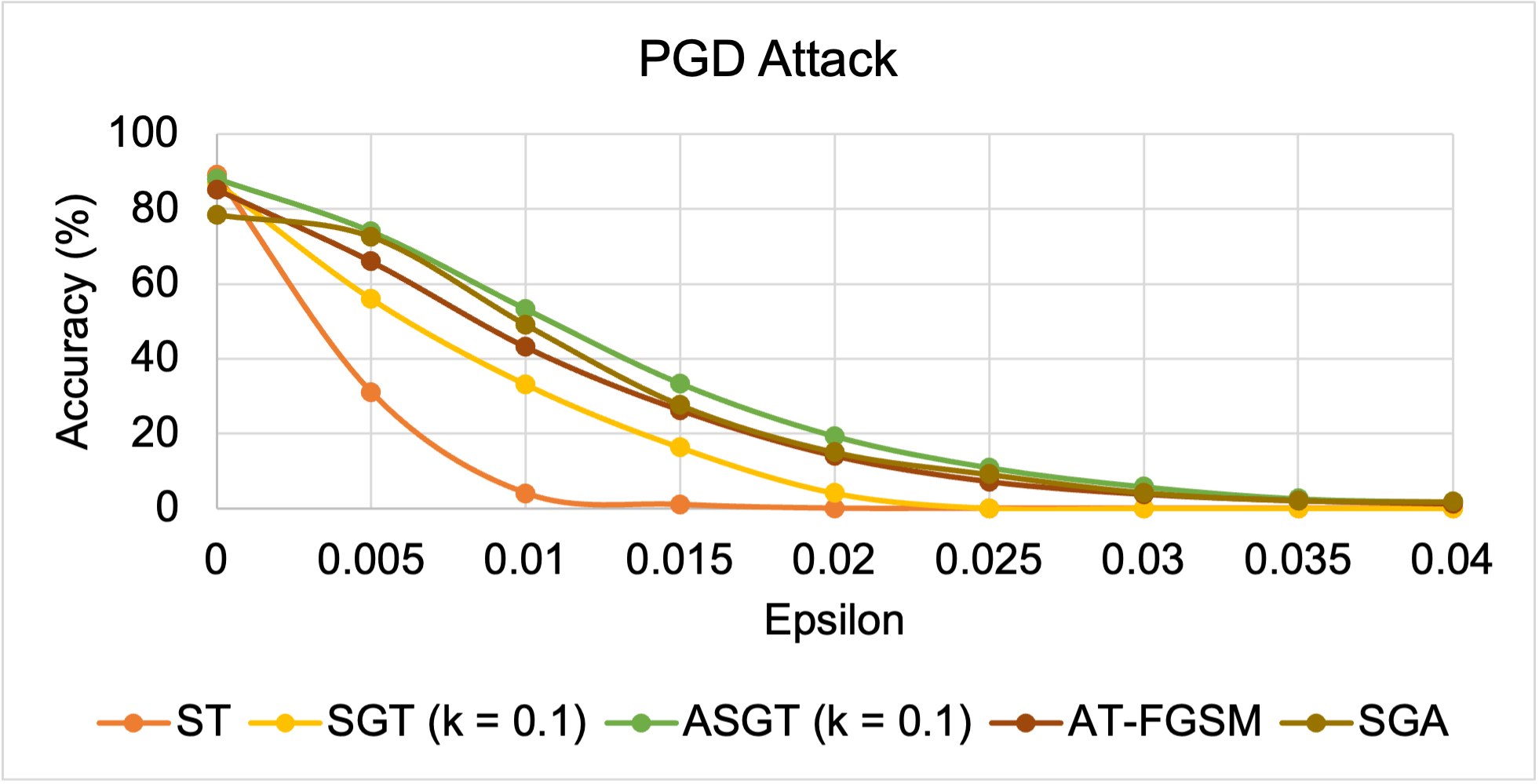}
\includegraphics[width=0.33\textwidth, height=3cm]{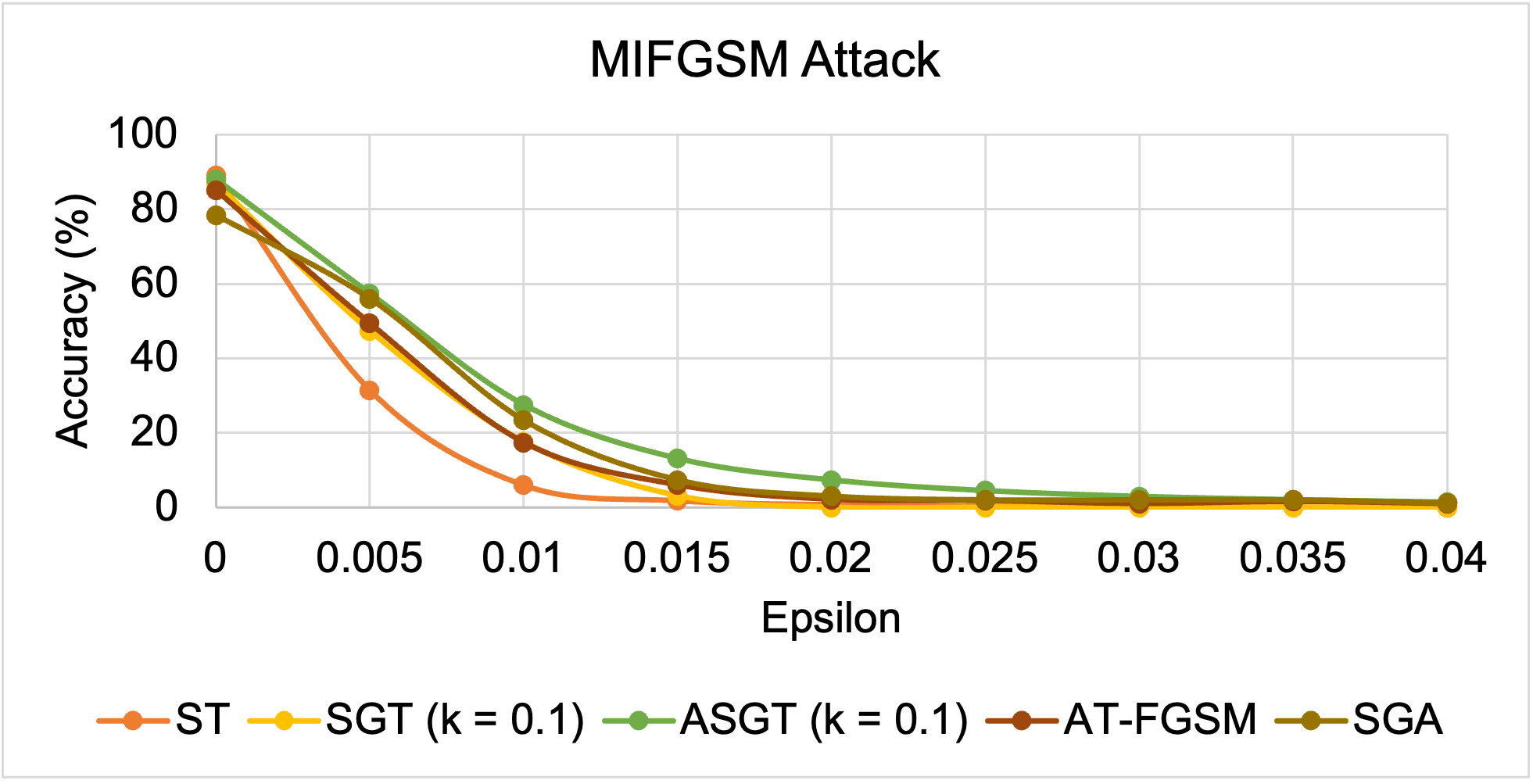}
\caption{Robustness of models against adversarial examples on the CIFAR-10 dataset. Models trained using ST, SGT (\( k = 0.1 \)), ASGT (\( k = 0.1 \)), AT-FGSM, and SGA across various magnitudes of noise (\( \epsilon \)) for the FGSM, PGD, and MIFGSM attacks.}
\label{fig:sgt_cifar_all}
\end{figure*}

\begin{figure*}[!tp]
    \centering
    \includegraphics[width=0.48\textwidth]{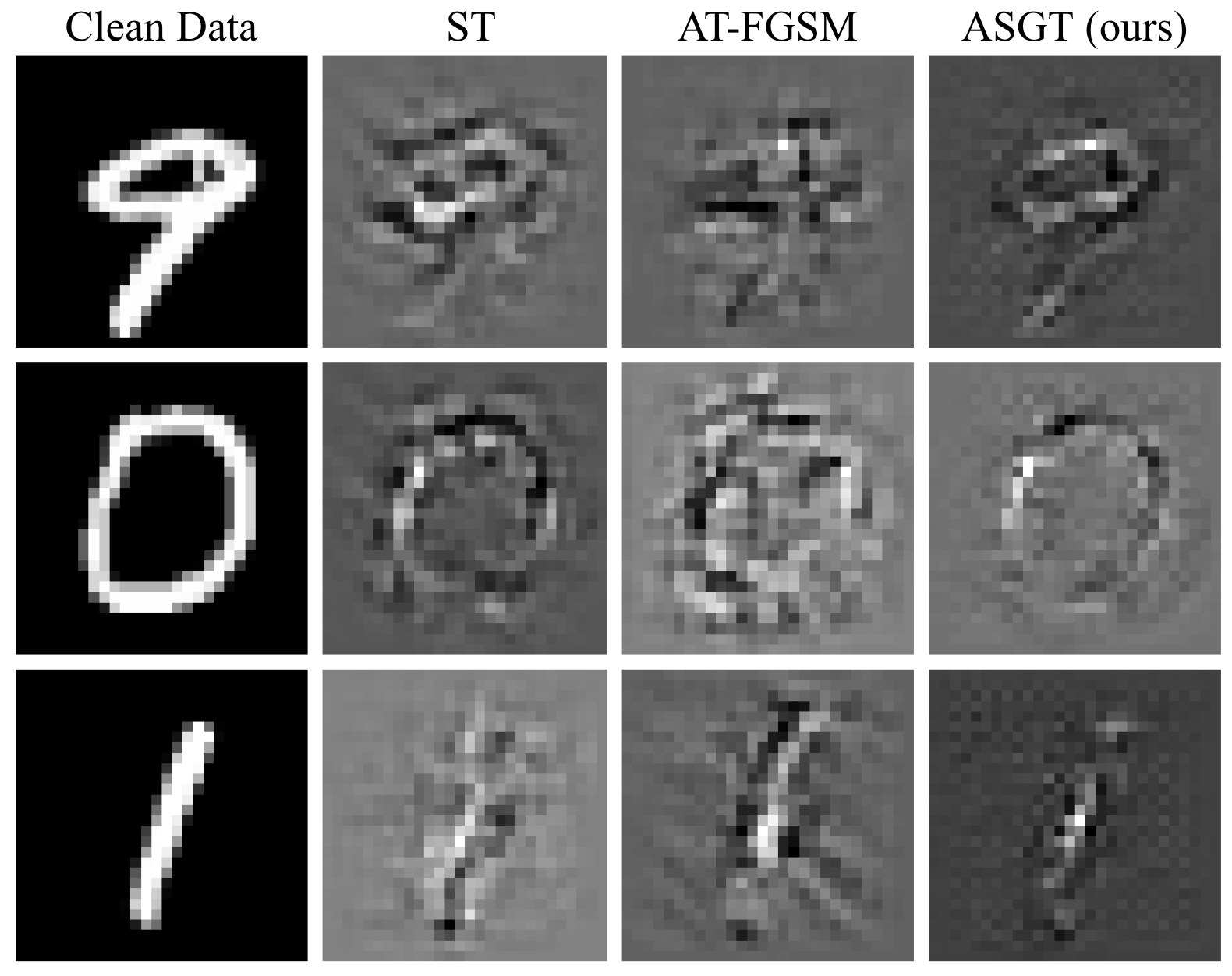}
    \includegraphics[width=0.48\textwidth]{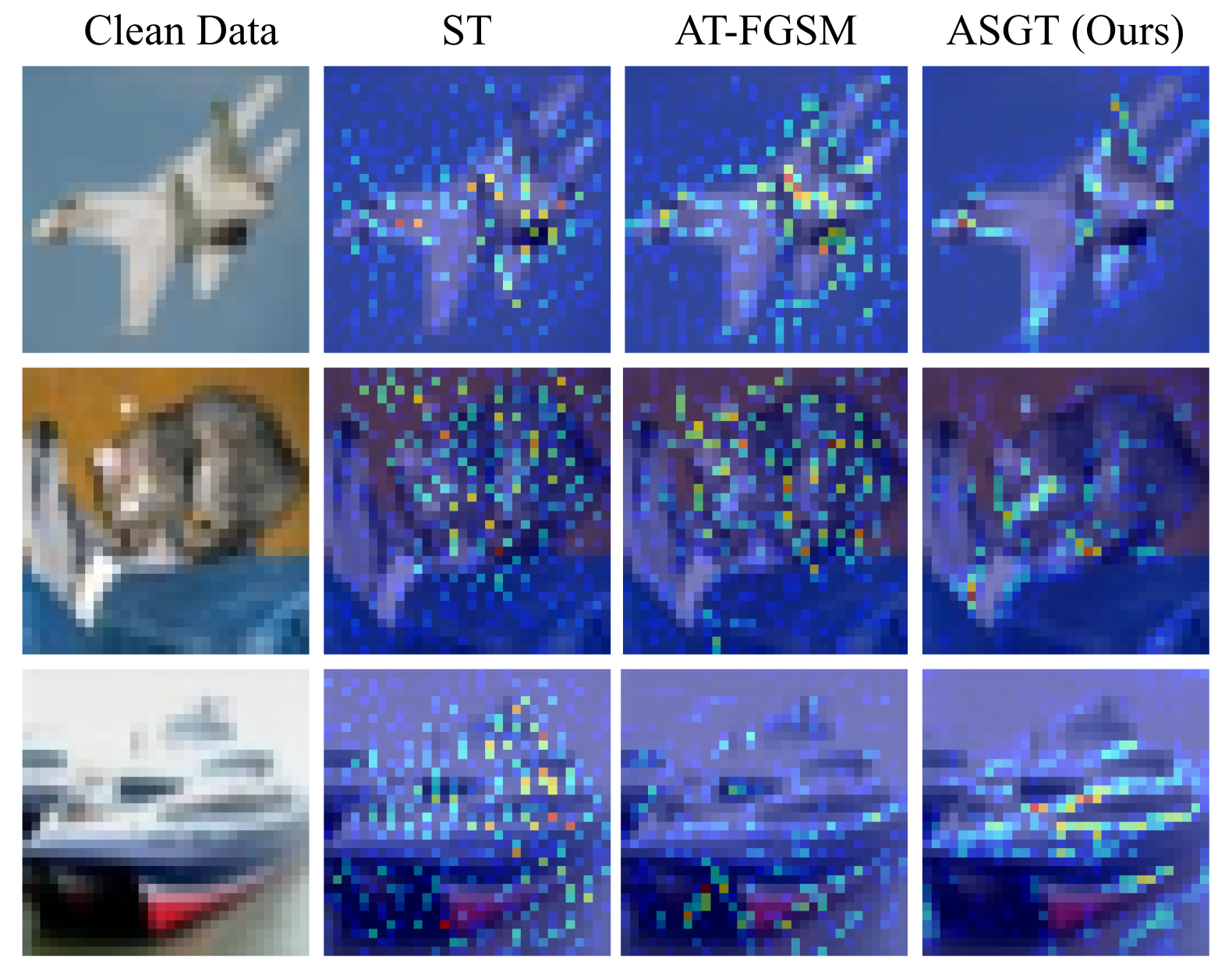}
    \caption{Saliency maps generated by standard training (ST), adversarial training (AT-FGSM), and adversarial saliency-guided training (ASGT). Left: MNIST dataset, Right: CIFAR-10 dataset.}
    \label{fig:interpretability}
\end{figure*}

\noindent \textbf{Datasets:} We conduct our experiments on MNIST \cite{mnist} and CIFAR-10 \cite{cifar} datasets. 

\noindent \textbf{Architectures:} We use AlexNet \cite{krizhevsky2017imagenet} architecture for MNIST dataset and ResNet-18 \cite{resnet18} for CIFAR-10 dataset. The models were trained using the Stochastic Gradient Descent (SGD) optimizer, with hyperparameters detailed in Table \ref{tab:hyperparam}. We note that the reported results represent the average of five independent tests.

\noindent \textbf{Threat model:} We operate under the assumption of a strong white-box scenario, where the attacker possesses complete knowledge of the victim deep neural network, including its architecture and parameters. 
The adversarial samples $X'$, used in the saliency-guided training, are generated by minimizing the distance, measured using the $L_{\infty}$ norm, between the model's output when fed the clean sample and the adversarial noise, with respect to the true label. The hyperparameter $k$ and $\lambda$ are fine-tuned as 0.1 and 1 respectively.

We evaluate the robustness of different trained models against three prominent adversarial attacks: Fast Gradient Sign Method (FGSM), Projected Gradient Descent (PGD), and Momentum Iterative Fast Gradient Sign Method (MIFGSM). 

\begin{table}
    \centering
    \begin{tabular}{ccccc}
    \midrule
        \textbf{Dataset} & \textbf{Architecture} & \textbf{Epochs} & \textbf{Batch size} & \textbf{Lr}\\
        \midrule
        MNIST & AlexNet & 20 & 128 & 1e-2\\
        CIFAR-10   & ResNet18 & 100  & 256 &  1e-4\\ 
        \midrule
    \end{tabular}
    \caption{List of hyperparameters used to train models.}
    \label{tab:hyperparam}
\end{table}

\noindent \textbf{Fast gradient sign method (FGSM)} \cite{fgsm} is a single-step, gradient-based, attack. An adversarial example is generated by performing a one step gradient update along the direction of the sign of gradient at each pixel as follows:

 \begin{equation}
     x^{adv} = x - \epsilon \cdot sign (\nabla_{x}J(x,y))
 \end{equation}
Where $\nabla J()$ computes the gradient of the loss function $J$ and $\theta$ is the set of model parameters. The $sign()$ denotes the sign function and $\epsilon$ is the perturbation magnitude. 

\noindent \textbf{Projected Gradient Descent (PGD)} \cite{at_madry} is an iterative variant of the FGSM where the adversarial example is generated as follows:
 \begin{equation}
x^{t+1} = \mathcal{P}_{\mathcal{S}_x}(x^t + \alpha \cdot sign (\nabla_{x}\mathcal{L}_{\theta}(x^t,y)) )
 \end{equation}
Where $\mathcal{P}_{\mathcal{S}_x}()$ is a projection operator projecting the input into the feasible region $\mathcal{S}_x$ and $\alpha$ is the added noise at each iteration. The PGD attack tries to find the perturbation that maximizes the loss of a model on a particular input while keeping the size of the perturbation smaller than a specified amount. 

\noindent \textbf{Momentum Iterative Fast Gradient Sign Method (MIFGSM)} \cite{dong2018boosting} introduced a momentum term to stabilize the update direction during the iteration.
 \begin{equation}
     g_{t+1} = \mu \cdot g_{t} + \frac{\nabla_{x}J(x_{t+1}^{adv},y)}{\parallel \nabla_{x}J(x_{t+1}^{adv},y) \parallel_1}
 \end{equation}
 \begin{equation}
     x_{t+1}^{adv} = x_{t}^{adv} - \alpha \cdot sign (g_{t+1})
 \end{equation}
 
\subsection{Does Saliency-Based Training Enhance Robustness for Deep Neural Networks?}

Our study aimed to investigate the influence of saliency guided training on the robustness of deep learning (DL) models. To accomplish this, we trained our models using: standard training (ST) and saliency-guided training (SGT), with varying degrees of feature masking denoted by parameters $k = 0.5$ and $k = 0.3$. Subsequently, we subjected the trained models to rigorous testing to evaluate their robustness against three prominent adversarial attacks: FGSM, PGD, and MIFGSM. 

For the MNIST dataset, we varied the magnitude of noise, \( Epsilon \), for each attack within the range of \( 0.05 \) to \( 0.4 \). As depicted in Figure \ref{fig:sgt_mnist}, the application of SGT demonstrated a notable improvement in model robustness against adversarial examples. Specifically, when \( Epsilon = 0.25 \), a model trained using standard training (ST) exhibited an accuracy of \( 27.88\% \). In contrast, models trained using SGT achieved significantly higher accuracies, with \( 51.79\% \) and \( 44.29\% \) for \( k = 0.3 \) and \( k = 0.5 \) respectively, when subjected to the FGSM attack.
A similar trend was observed when the models were exposed to PGD and MIFGSM attacks, with gains of \( 20\% \) and \( 30\% \), respectively, at \( Epsilon = 0.25 \). 

Using SGT on the CIFAR-10 dataset led to a notable enhancement in the model's resilience against adversarial examples. Our experiments across multiple attack scenarios revealed a substantial improvement in model accuracy, as illustrated in Figure \ref{fig:sgt_cifar}. Particularly noteworthy was the model's performance when subjected to the Fast Gradient Sign Method (FGSM) attack with a noise magnitude of $\epsilon = 0.025$, where it demonstrated an impressive accuracy gain of over $20\%$. Similar improvements in accuracy were consistently observed across all other tested attack methods.

These findings underscore the efficacy of SGT in bolstering model resilience against adversarial perturbations across a range of attack scenarios.




\subsection{Impact of Adversarial Saliency-guided Training on both Model Robustness and Interpretability}

\noindent \textbf{Impact on Robustness:}
We compare various training methods, including ST, SGT, AT, SGA \cite{li2022saliency}, and ASGT, in terms of resulting model robustness.
Models trained on the MNIST dataset using ASGT exhibited the highest robustness across different attacks. For instance, under the PGD attack with $\epsilon = 0.2$, the accuracy of the ST model was $16.36\%$, $49.08\%$ for SGT, $8.33\%$ for AT-FGSM, $48.95\%$ for SGA, and $55.69\%$ for ASGT (see Figure \ref{fig:sgt_mnist_all}).

Training the model on the CIFAR-10 dataset using Adversarial Saliency Guided Training (ASGT) noticeably improved its robustness against adversarial examples. Across various attacks, we observed a significant increase in model accuracy, as depicted in Figure \ref{fig:sgt_cifar_all}. Notably, when subjected to the Fast Gradient Sign Method (FGSM) attack with a noise magnitude of $\epsilon = 0.025$, the model exhibited an impressive gain of over $30\%$ in accuracy. Similar enhancements in accuracy were observed across all other attacks tested.

\noindent \textbf{Impact on Interpretability:}
Alongside the improvement in robustness, we also achieve enhanced interpretability. Figure \ref{fig:interpretability} illustrates the heat maps generated by models trained using various training methods. ST yields the least satisfactory heat map interpretations. AT slightly outperforms ST but still exhibits some presence of non-relevant features. In contrast, ASGT consistently produces superior heat maps, focusing predominantly on the target object with fewer irrelevant features present.

\section{Conclusion}
\label{sec:conclusion}

In this study, we delved into the intricate relationship between interpretability and robustness in deep learning models, focusing on the impact of Saliency-guided Training (SGT) and proposing a novel combined training technique. Our investigation revealed that SGT indeed enhances model resilience against white-box adversarial attacks, contrary to previous findings. Moreover, through the integration of SGT and adversarial training (AT), we introduced a novel training approach aimed at producing deep neural networks that are both robust and interpretable.

\vfill\pagebreak

\bibliographystyle{IEEEbib}
\bibliography{refs}
\balance
\end{document}